# Online People Tracking and Identification with RFID and Kinect

Xinyu Li, Yanyi Zhang, Ivan Marsic, and Randall S. Burd

*Abstract*—We introduce a novel, accurate and practical system for real-time people tracking and identification. We used a Kinect V2 sensor for tracking that generates a body skeleton for up to six people in the view. We perform identification using both Kinect and passive RFID, by first measuring the velocity vector of person's skeleton and of their RFID tag using the position of the RFID reader antennas as reference points and then finding the best match between skeletons and tags. We introduce a method for synchronizing Kinect data, which is captured regularly, with irregular or missing RFID data readouts. Our experiments show centimeter-level people tracking resolution with 80% average identification accuracy for up to six people in indoor environments, which meets the needs of many applications. Our system can preserve user privacy and work with different lighting.

## I. INTRODUCTION

Real-time people tracking and synchronized identification are needed in various application, such as monitoring the work of team members with different roles, tracking persons of interest (an elderly or a patient), while ignoring others (e.g. visitors) or establish the connection between players and virtual reality roles based on the tags the players wearing. Tracking and identification are often treated as two separate tasks and addressed using different methods. Previous work has used cameras [1][2][3] or robots with different sensors [4] to track multiple people. For vision-based approaches, usually a clear facial image is required for face recognition and people identification. This approach has limitations for real-world applications as people may not always face the camera, or face it at different angles, and appear under different light conditions. Identification based on face recognition also raises privacy concerns in some settings. We present an approach for tracking and identifying people in real time that does not need to compromise privacy. Such approach is based on sensory data that do not directly reveal people identities. We use depth imaging and passive RFID for tracking individuals in real-world coordinates and retrieving some aspect of their identity (code name or role in a team) from a lookup table based on ID numbers of RFID tags.

Previous studies have used approaches that first locate people in RGB video frames and then obtain their identity using facial recognition. We use depth sensing instead because it is not influenced by light conditions, and the depth information allows accurate projection of depth-frame pixels into the real-world. Kinect, a commercial depth sensor, includes a software development kit (SDK) that makes people tracking in real-time a manageable problem. However, even with a depth sensor, real-time people recognition remains a challenge. Limitations include low resolution of depth sensors, limited computational power for real-time recognition, accuracy of facial recognition algorithms, and the need for face visibility in the camera view.

In addition to computer vision, mobile sensors, such as passive RFID, have been used for people identification based on the ID number of the tag they are wearing. The challenge is that it is very difficult to accurately track moving people based on passive RFID alone. Several systems have been proposed that use passive or active RFID for people and object tracking [5][6][7][8] but they either show low tracking accuracy or work only in relatively slow-moving environments.

We designed the system to take the advantage of Kinect depth sensor and used a rotation matrix to project each person's head joint into the floor plane for people tracking. Inspired by the previous research which has shown that people's walking speed is influenced by their age, weight and health conditions, so that the moving speed varies from person to person [9], we use each person's moving speed and direction for identification. We measured the velocity of people in the room using Kinect, and the velocity of passive RFID tags attached to people [10], and matched the tags to people based on their velocities. Most previous research based on phase angle used RFID readers at a constant working frequency [8][10]. However, frequency hopping influences the measured phase angle since phase angle is estimated based on carrier frequency. To deal with this problem, we introduced a method that ensures accurate RFID-based velocity measurement of tags even when the RFID reader uses frequency hopping. Our system is able to measure the velocity at a centimeter level with both Kinect depth sensor and RFID by using Doppler frequency shift estimated from phase angle changes. Unlike Kinect, where depth-images are recorded at a regular rate, the RFID reader uses a communication protocol that results in unpredictable intervals between successful measurements of phase angles. To address this challenge, we propose the "drop" method which enables the system to match the tags to people by calculating the distance matrix between their velocity sequences. The contributions of the paper are:
1. A novel, practical and inexpensive framework for multiple people tracking and identification that works in real-time.
2. A strategy for Doppler frequency estimation using phase angle that works under RFID-reader frequency hopping.
3. A practical strategy to manage missing data points when matching the velocities measured by different sources without damaging the sequence synchronization.

The rest of this paper is organized as follows. Section 2 introduces related research. Section 3 describes our approach to people tracking and identification. Section 4 presents experimental results of our system. Section 5 discusses the system limitations and potential future improvements. Finally, Section 6 concludes the paper.

## II. RELATED WORK

People tracking has been researched for decades with computer vision being the most common approach. Initial research used RGB video frames and several local and temporal features [1][2][3]. These approaches, however, only locate people in the video frames, and do not perform people tracking in a 3D environment or estimate people's location in real world coordinates. To quantify a person's coordinates and motion, multi-camera based solutions [12] and depth sensing have been researched in recent years [13][14]. A prototype system for this approach has not been widely implemented due to limited tracking accuracy, cost and privacy concerns. Recent development of commercial depth sensors has made possible multiple people tracking with a few lines of code. For example, the Microsoft Kinect provides the 25-joint-skeletons of up to 6 people in the camera space at about 30 frames per second. Although people tracking using commercial depth sensors or RGB-D cameras has become a manageable problem, computer-vision-based people identification still faces several challenges. A key challenge is that in dynamic and complex environments people may be occluded and often do not face the camera or assume different postures. In addition, people identification from facial images captured under various lighting conditions, resolutions and view angles is still inadequate [15][16].

Compared to vision-based systems, people identification systems based on mobile sensors and passive RFID have advantages because each sensor has a unique or programmable ID, which can be used for identification [17]. For mobile sensors like gyroscopes or accelerometers [18], tracking moving speed and direction requires additional information to determine person's coordinates. This approach also suffers from a cumulative error which requires additional information to calibrate the sensors. Recent studies have shown that people tracking and identification can be accomplished using RFID [5][6][7][8]. The RFID-based solution is suitable for real-world applications because passive RFID tags are small, cheap, and battery-free, hence easily attached to people or objects. Three RFID-based tracking strategies use antenna detection zones, received signal strength (RSS), and the angle of arrival of RFID signal. The first strategy identifies the zone covered by the antenna that detected the tag, also known as the "zoning" method [2]. This method determines whether a particular tag is within one or more coverage areas. Although easy to implement, this method suffers from low tracking resolution, depending on the shape of overlapped antenna coverages (varies from decimeter to meters). The second strategy estimates the exact position of RFID tags using RSS. Several approaches then use a reference-tag method that compares the RF signal between target tags and reference tags [6][7][19]. People tracking using reference tags, however, is not as accurate as camera-based tracking and the tracking resolution is usually at the meter level. In addition, placing reference tags in the environment is often inconvenient and obtrusive. The third strategy has used the phase angle of RFID signal arrival to measure the position of the tag [8][11]. Results show promising performance for tracking in an experimental environment, achieving up to centimeter-level accuracy [8]. However, this approach requires a known starting point to determine the coordinate of the tag's position, a relatively controlled environment, and the RFID reader to work at a fixed frequency band to minimize the sources of noise, which limits the practicality of such systems.

Several approaches to simultaneous tracking people and retrieving their identities combine the data of multiple sensors. A robot with RFID antennas and a camera was used for people tracking in crowded environment [21]. However, the robot could not keep tracking multiple people walking in opposite directions since it had only one camera installed. In addition, in many applications a roaming robot may interfere with people's work. A further refinement of this approach used a fixed camera and RFID antennas for people tracking and identification [22][23]. The system was easy to implement in real-world applications, but the RFID positioning method (zoning) is not accurate and cannot distinguish two people standing close to each other. Estimating of people in the room with a single RGB camera and project their location into 2D space is not accurate [23]. Our proposed approach for people tracking and identification achieves high resolution and accuracy using depth information and RFID data captured by a commercial camera and fixed RFID antennas. Our approach does not attempt identification by matching people's locations estimated from RFID and camera view [23]. Instead, our system tracks people using a depth camera and performs identification by matching the relative moving velocities of people (from depth sensor) with velocities of the tags they are wearing (from fixed RFID antennas). The system was tested in a similar indoor environment with multiple people moving or standing still as previous research did [21][22][23] and easily achieved centimeter-level people tracking resolution. We also achieved the comparable identification accuracy compared with systems with meter level tracking resolution [22][23].

## III. APPROACH

We refer to people tracking as quantifying person's position in 2D coordinates of a room layout map. Layout mapping can be done with a single Kinect sensor [24], where each pixel in the room layout map represents $1cm^2$ in the real world. We refer to people identification as retrieving an aspect of person's identity from a lookup table, so that the identity can be the name, or the role of the person in a team. Because of the Kinect limitation, our system can track and identify up to 6 people in the area of interest. To cover a larger area with more people, more Kinect can be installed as [25].

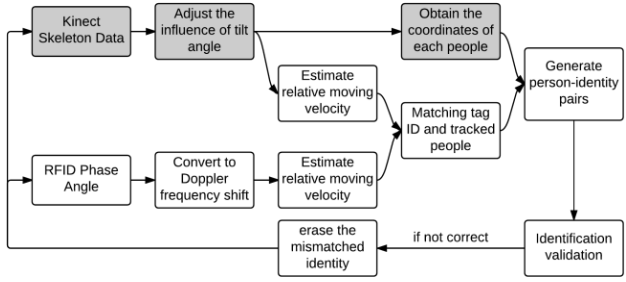

Fig. 1. Components of our tracking and identification system. Shaded components are for people tracking and unshaded ones are for identification.

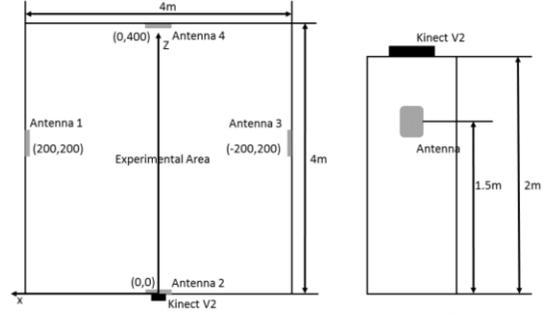

Fig. 2. Left: Floor plane of our hardware configuration with the coordinate axes *X* and *Z* indicated. Right: Side view of Kinect on a cabinet 2m above the ground and two RFID antennas on the wall 1.5m above the ground.

Our system accomplishes tracking and identification with two subsystems (Fig. 1): a people tracking subsystem and people identification subsystem. The people tracking subsystem takes each person's head joint provided by Kinect and projects it into the floor plane using a rotation matrix (Fig. 1, shaded) and keeps estimating the moving velocity of each person. The people identification subsystem matches the velocity vectors of passive RFID tags and people. The system assigns the tag IDs to people based on the best match of velocity vectors. The system continually checks if the person's identity remains consistent over time and corrects any identification errors (Fig. 1, unshaded bottom feedback loop).

The following sections first introduce our method for measuring the relative velocities of people and RFID tags using Kinect depth sensor and passive RFID. We then present our approach to identification by matching the tags to people.

*A. System Overview*

We used commercially available devices to increase the applicability of our system. For people tracking we used the depth sensor in the Microsoft Kinect V2. To achieve a better view of the room and avoid occlusions, we mounted the Kinect sensor two meters above the ground and tilted the camera downwards with a tilt angle $\theta$, $\theta=10°$ in this paper (Fig. 2). We used a similar configuration previously and achieved adequate performance in a similar environment [24]. The Impinj R420 RFID reader was used with Alien 9611 antennas to measure the signal from passive RFID tags. We mounted the RFID antennas facing horizontally at 1.5 meters above the ground, which is the height of a typical person's chest (Fig. 2). Note that mounting the antennas on the ceiling or putting the antenna not at roughly same height as the tag will change the velocity measurement from 2D to 3D, but the concept stays the same.

*B. People Tracking Based on Kinect*

*1) People tracking in real-world coordinates*

We choose the Kinect depth sensor over RGB imaging for two reasons. First, the depth sensor works independent of light conditions which allows our system to work at night or under variable light conditions. For example, in laboratories or medical settings an extra light source may be used for certain applications, or the lights may be switched off to view x-ray images. Second, the depth data from the Kinect SDK provides the coordinates of people's joints in Kinect camera space (the coordinate system defined by the Kinect SDK). In camera space, the *X*-axis grows to left of the sensor, the *Y*-axis grows up and the *Z*-axis grows out in the direction the sensor is facing, and each pixel corresponds to 1 cm in the real world. These parameters make it possible to determine a person's location.

People's foot coordinates cannot be directly used as their position, because the feet are often occluded by other people or objects in the room, resulting in inaccurate and unstable coordinate estimates [24]. Instead, we used the projection of the head to the floor plane as person's position, because head is the least likely occluded body part. Our Kinect positioning (Fig. 2) ensures the heads are maximally exposed to the camera. The remaining problem is that the Kinect camera space is not parallel with the world coordinate system due to the tilt angle. This problem is solved using the rotation matrix:

$$[x', z'] = [x, y, z] \cdot \begin{bmatrix} 1 & 0 \\ 0 & -\sin(\theta) \\ 0 & \cos(\theta) \end{bmatrix} \quad (1)$$

where (*x*, *y*, *z*) is the coordinate of head joint in Kinect space, (*x'*, *z'*) is the projected head coordinate and $\theta$ is the tilt angle decided when the Kinect is installed. We use the rotated coordinates as the person's position, and distinguish different people using the skeleton IDs provided by the Kinect SDK.

*2) Estimating people's relative velocity*

Given the people's coordinates in the room, their velocities can be represented using the velocity components on X-axis and Z-axis (Fig. 2). Because RFID cannot accurately measure the tag's movement in the floor-plane coordinates, we radial coordinates for both people and tags. We measured the radial velocity of each person in the floor plane with the projected location of the RFID antennas as the reference points:

$$V_{person} = \frac{\Delta D}{\Delta t} \quad (2)$$

where $\Delta D$ is the moving distance relative to the reference point and $\Delta t$ is the time interval between subsequent position measurements. We defined the velocity as positive if a person is moving to the reference point or negative if a person is moving away from the reference point.

*C. RFID Tag Tracking*

*1) Estimating Doppler frequency shift*

Similar to people velocities, we estimated the velocity of each tag relative to each of RFID antennas using Doppler frequency shift (Doppler shift). Current RFID readers have a built-in function to measure the Doppler shift, but this measurement is not accurate enough for many applications [10]. We observed that the Impinj R420 reader measures the Doppler shift with a standard deviation of 2.68 rad/s even when the tag is stationary. This large deviation leads to 44cm/s to 49cm/s deviation in tag velocity measurement, which is quite imprecise compared to the measurements provided by Kinect [27]. Therefore, instead of reading the Doppler shift directly from the reader, we calculated the Doppler shift $f_D$ using the phase angle changes measured by the reader [10]:

$$f_D = \frac{1}{4\pi} \cdot \frac{\Delta\varphi}{\Delta t} \quad (3)$$

where $\Delta\varphi$ is the phase angle difference between the transmitted and the received signal, and $\Delta t$ is the duration of signal transmission. Placing tags on people's chest makes the tag roughly at the same height as the reader antenna, so the problem of velocity estimation can be considered in 2D-plane only. Previous research on measuring the phase angle changes and estimating Doppler shift operated the RFID reader at a fixed carrier frequency to avoid deterioration in the phase angle measurement caused by RFID frequency hopping [8][10]. However, using RFID reader with a fixed frequency may cause signal interference to other devices that share the same band and is not allowed in many countries, including the United States. We operated the RFID reader in the "*MaxMiller*" reading mode to achieve maximum reading speed and try to provide sufficient number of phase measurements for velocity estimation before frequency hopping occurs. We used only the pairs of phase angles measured at the same carrier frequency to calculate the Doppler shift (Eq. (3)). The accuracy of Doppler shift estimation depends on the length of the interval $\Delta t$ between two phase measurements because the phase angle provided by the RFID reader is modulo of $2\pi$. We can make the estimation accurately within every 400ms because the phase change $\Delta\varphi$ will not exceed $2\pi$ within 400ms. If $\Delta t$ gets larger, $\Delta\varphi$ might be greater than $2\pi$ for an unknown number of $2\pi$ cycles. However, the measured phase angle returned by reader will be mod of $2\pi$, thus preventing the exact calculation of phase change $\Delta\varphi$.

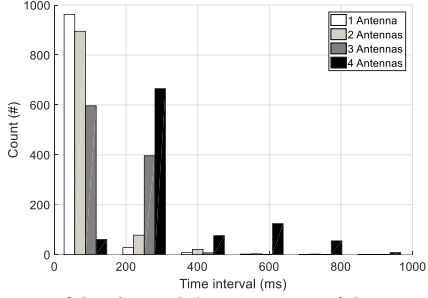

Fig. 3. Histograms of time intervals between successful measurements of phase at any carrier frequency for different number of reader antennas and 25 tags. 1000 measurements were performed for each figure.

When the frequency hopping occurs, our system drops the last phase measurement because it cannot be paired with the next measurement at the new carrier frequency and starts to measure the phase angle in the new carrier frequency band. Based on our experiments, the Doppler shift measured by our method under frequency hopping has only a 0.18 rad/s standard deviation when the tag is still (with 25 tags in the room and 2 reader antennas), which is almost 15 times less than the Doppler shift provided by the API. The time interval between two successful phase measurements for same tag by same antenna (given same reading mode and same number of tags presented in the environment) depends on the number of reader antennas since antenna switching takes time. The time interval between two successful measurements will be longer with more antennas attached to reader since the RFID reader has to loop through all the attached antennas within 400ms to accomplish a successful velocity measurement (Fig. 3).

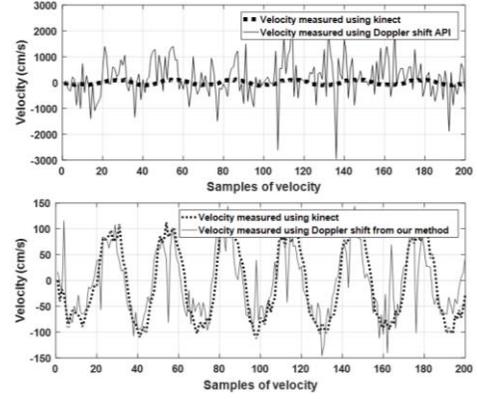

Fig. 4. Comparison of tag velocity measured using Doppler shift from RFID reader (Top) and our method (Bottom). The Kinect measurement served as the ground truth in both cases. Note the difference in vertical scales. Each sample takes 0.2 seconds.

*2) Estimating relative velocity of tags*

Given the Doppler frequency shift, the moving speed of the tag can be estimated using the following equation:

$$v = \frac{c \cdot f_D}{2 \cdot f_t} \quad (4)$$

where $c$ is the speed of the light and the $f_t$ is the transmitter's frequency, which is generated by the RFID antenna and is known to the RFID reader. Since the estimated Doppler frequency shift will be negative if the tag is moving further to the antenna and positive if the tag is moving closer to the antenna, the estimated relative velocity follows the same definition as used above for people tracking with the Kinect.

We compared the relative velocity estimation using the Doppler shift provided by reader and by our method. We had 10 volunteers participate in the experiments with a tag on a name badge worn on their chest. To avoid the noise caused by the movement of several people, participants moved one at a time in the room. We used the relative moving velocity measured by the Kinect as the ground truth because Kinect can achieve accuracy of 8 to 9 cm for tracking head joints [27]. Our experimental results showed that the velocity to antenna 1 estimated from the Doppler shift provided by the reader had an average error of 280cm/s (Fig. 4 top) . The velocity to antenna 1 estimated using our Doppler shift method has around an average error of 23cm/s, which is 12 times smaller than the error using the Doppler shift measured by the reader (Fig. 4 bottom).

### D. Matching Tags to People

*1) Drop method*

Because the number of successful phase readings (or, read rate) is not constant for passive RFID, and the rate of depth frames from Kinect slightly varies for every second, we cannot directly compare and match the relative velocity of the RFID tag to a moving person. To address the mismatch of data rates, we programmed both RFID and Kinect systems to measure the relative velocity of each tag and person every 400ms. We choose 400ms since in the United States, RFID reader frequency hopping is regulated to occur within 400ms, so having the window size larger than 400 seconds would not improve phase measurement. The normal frame rate for Kinect

| Dropped data | | | | | | | | | | |
|---|---|---|---|---|---|---|---|---|---|---|
| Velocity of person 1 | X | X | X | X | X | X | X | X | X | X |
| Velocity of person 2 | X | X | X | X | X | X | X | X | X | X |
| Velocity of person 3 | X | X | X | X | X | X | X | X | X | X |
| Velocity of tag 1 | X | X | O | X | X | X | X | X | X | X |
| Velocity of tag 2 | X | X | X | X | X | O | X | X | X | X |

X: Valid velocity measurement
O: Failed velocity measurement

Fig. 5. Illustration of the "drop" method. If one of the RFID antenna fails to measure the velocity of a tag, all the velocity measurements at that time instance will be dropped (marked with dashed line rectangles). The data was collected with 3 people in the room, two of them had tags on them.

is 30 frames per second so it can easily measure the relative velocities of people every 400ms.

Although we set the RFID reader at the maximum read speed to try to ensure the phase angle of each tag is estimated at least twice before frequency hopping happens, the system may still not succeed. With fewer than two phase angle measurement, the system cannot estimate the velocity at a given time point. As a result, there may be gaps in the velocity sequence of tags, which then obstructs the time synchronization when matching the velocity sequences of tags and people.

A common approach to compare two similar time series is dynamic time warping (DTW) [28]. However, the RFID-based velocity measurement for a stationary tag, unlike the Kinect-based velocity of the person wearing the tag, is not a sequence of zeroes due to the noise in the RFID system. The DTW finds the shortest distance between two sequences, which leads to a high matching distance between an all-zero sequence and a sequence with a zero mean but non-zero standard deviation. To address this problem, we introduce a "drop" method (Fig. 5). When one of the reader antennas fails to collect enough phase-angle readings from a tag to calculate its velocity, our system drops all other velocity measurements made by RFID and Kinect at that time point.

We found that the number of dropped measurements was influenced by the number of tags in the room and the number of antennas attached to each RFID reader. If we define the drop rate as the percentage of data points been dropped before 20 sets of synchronized data points was collected, our experiments show that the system is able to work with less than 10% drop rate with 50 tags in the room using single antenna and 25 tags in the room using 2 antennas (Fig. 6). More antennas can be used to add additional dimensions to the velocity measurement to improve the matching accuracy, however, since the phase angle changes must be measured very fast, based on our experiments (Fig. 6), at most two antenna has to be attached to a different RFID reader. The compromise here is between accuracy vs. cost. Since the RFID reader rarely fails to estimate the moving velocity, we decide that if a tag is not providing enough phase readings for three contiguous measurements, we consider that this tag left the area of interest and exclude it from the people identification process. If the tag reappears, its tracking will be resumed. Compared to using smooth filtering or interpolation to fill the sampling gaps caused by missing data, the proposed "drop" method has two advantages. First, velocity measured during different time interval and by different sensors are independent with each other, therefore, dropped data points do not introduce extra errors into the system, while using smooth filtering would change the measured velocity values and interpolation will add extra value to the sequence, in a sense adding "noise" to original measured data. Second, all instances where velocity was measured were synchronized in time and interpolation or filtering may compromise the synchronization of these sequences.

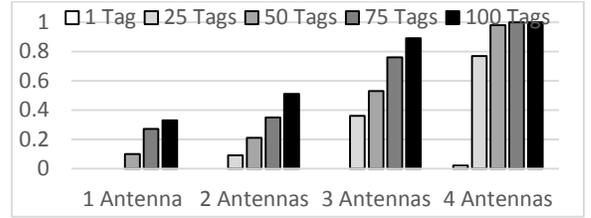

Fig. 6. Drop rate of the system with different of antennas attached when different number of tags in the rooms.

*2) People identification*

People identification is performed by finding the best-matched person for each of the RFID tags in the area using the velocity sequence of a given person and tag, then assigning the identity based on the tag ID to the matched person. In practice, we calculate the distance matrix of tag velocity sequences and people velocity sequences using every 20 data points after a drop, this ensures that all the velocity measurements are valid and synchronized in time. We simply define the distance matrix as the Euclidean distance for calculation simplicity.

Fig. 7 gives an example of calculating the distance matrix of two tags with two people. We match each tag ID to the person whose velocity sequence has the minimum distance compared to the tag's velocity sequence. From the figure, person 1 matches tag 2, person 2 matches tag 1 and the distance matrix also confirms these observations.

Once an identity is assigned to a person, normally the velocity sequence of the person will not be used for the distance matrix calculation. The two exceptions are the person leaves the camera view area and the identity is considered to be incorrect and canceled by the system, as will be introduced next.

*3) Identification validation*

We designed the system to work for real-world application scenarios but there are certain scenarios where this system will yield more inaccuracies. For example, two people are standing still, the measured velocity of people and tags will be very similar to each other and the system may assign the wrong identity.

To address this problem, besides the identification loop where the system continually uses a 20-point velocity sequence of each person and tag to calculate the distance matrix, we designed a validation loop to cancel incorrectly signed identities. The validation is performed by calculating the average distance of each pair of matched velocity sequences (tags and people) using a time window. We use a 40-sample window in this paper for the balance of validation time and accuracy (larger sample window will lead to longer validation time but more reliable validation performance). A threshold is used to determine whether the identification is valid or not; if the average distance between the tag velocity sequence and person velocity sequence exceeded the threshold, the system will de-match the tag and person and put them into the identification loop. The threshold is determined based on the average distance between the velocity sequences of tags and people from our experiments. In our application environment, the average distance is around 23cm/s based on 10,000 data points we collected with 10 different people. We set the threshold to be 30 which is slightly higher than the average distance to ensure the distance is not caused by the error of measurement.

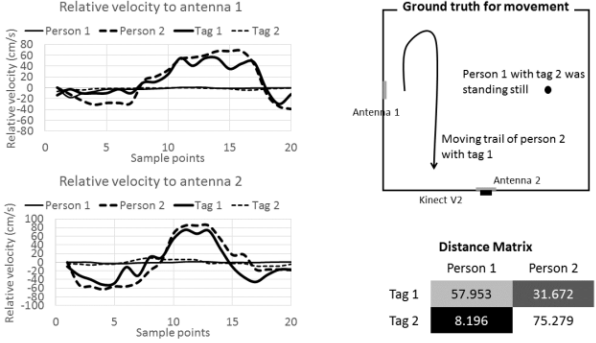

Fig. 7. Left: the estimated relative velocity of people and tags to antenna 1 and antenna 2 measured by Kinect and RFID. Right: the movement ground truth and calculated distance matrix. Lower distance indicates better match.

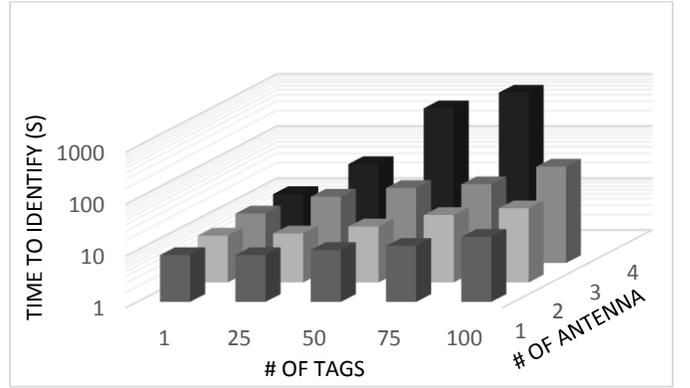

Fig. 8. Average time required for each successful identification with different number of tag s present for single person.

## IV. EXPERIMENTAL RESULTS

### A. People Tracking

We first evaluate the people tracking accuracy of our system and compare our system performance with other well-known systems. To accurately measure the people tracking accuracy, we put the markers on the floor of the experimental area and ask people to walk along the path we marked. The position of each person is recorded by our system and used for evaluation. We performed 100 experiment runs in total with 10 volunteers (5 males, 5 females) with 5 different moving paths. As it is mentioned, the system continually estimates the person's location coordinate in a 2D plane and each pixel in the plane represents 1 cm in the real world. This enables us to evaluate people tracking performance by comparing the recorded coordinates with the moving path we used.

The experimental results show the root-mean-square error (RMSE) for 100 experiments is 18.2cm. We compared our results to selected research that used either camera or RFID for people tracking [6][7][8][20][27] (Table I). As it is shown in the table that our system utilized the advantage of depth sensor for people tracking and outperformed other systems. Note that the Tagoram [8] system, which is one of the best RFID-based tracking systems, achieved centimeter level accuracy in object tracking. When performing people tracking, since each person's moving trail is not pre-known and each people tracking in real applications (with frequency hopping and people moving) will take the people tracking accuracy to decimeter level.

### B. People Identification

For people identification, we tested our ability to identify individuals with one, two, four and six people in the experimental area (Fig. 2, 16m2 in this paper). We stopped at the scenario with six people due to the limitation of experimental space and hardware. Two aspect of the people identification was evaluated: the identification speed and the identification accuracy.

To test the identification speed with different number of RFID tags in the room, since the we asked a single person wearing tag to walk in and out of the experimental area and room and keep tracking the time it takes for system to identify correctly recognize the person. The results (Fig. 8)) show that the time takes to identify single person increase with the number of tags in the room and the number of antennas. Though theoretically, more antenna will lead to better identification accuracy, it will also lead to longer the time it takes for identification time. Our experimental results in actual application should be considered and we suggest to use two antennas with less than 50 tags to achieve around 10 seconds identification time can be achieved with 2 antennas under 50 tags scenario which should fit into most applications.

To evaluate the people identification accuracy, we programmed the system to record the identification results every second and calculate identification accuracy as:

$$accuracy = \frac{\# \ of \ people \ correctly \ recognized}{\# \ of \ people} \quad (4)$$

We performed several iteration of the experiments with different number of people and different people combination, each iteration lasts for 5 minutes. We evaluate the people identification accuracy using min, max, average accuracy and standard deviation (Table II). The experiments show that system has perfect performance when there is only one person or two people in the room. The performance gets worse when there are four people in the room because people movement cause signal reflection and further introduce extra noise to the phase measurement. Besides, with more people in the room, it is more likely that two people share the very similar moving pattern and lead to identification errors.

TABLE I
THE COMPARISON OF DIFFERENT TRACKING SYSTEMS.

| System | Sensor Type | Reported Accuracy |
| --- | --- | --- |
| LANDMARC [6] | Active RFID | ~2 meters |
| Tagoram [8] | Passive RFID | 20cm under 90% CDF for object tracking |
| Non-intrusive localization [7] | Passive RFID | 80cm under 90% CDF for object tracking |
| People tracking in sport [26] | Dual RGB cameras | 20-50cm |
| Our system | Kinect depth sensor | 19cm for a single person present |

TABLE II
THE PEOPLE IDENTIFICATION ACCURACY / SINGLE PEOPLE TRACKING ACCURACY WITH DIFFERENT NUMBER OF PEOPLE IN THE ROOM.

| # of people in the room | 1 person | 2 people | 4 people | 6 people |
| --- | --- | --- | --- | --- |
| Min accuracy | 1/n | 0/0 | 0.5/0.5 | 0.33/0.33 |
| Max accuracy | 1/n | 1/1 | 1/1 | 1/1 |
| Average accuracy | 1/n | 0.83/0.85 | 0.82/0.83 | 0.62/0.71 |
| Standard deviation | 0/n | 0.172/0.132 | 0.318/0.295 | 0.357/0.309 |

There is not much previous research about simultaneous people tracking and identification. We first compared with pioneer research using data fusion for people tracking and identification [22]. Our system achieved better tracking accuracy and identification accuracy since we are tracking people with a depth sensor which provides us with high accuracy and we do not rely on RSS for people identification. Recent research using computer vision and RFID for people tracking and identification achieved meter-level people tracking accuracy [21] whereas our system achieved centimeter level of people tracking accuracy in similar environment.

*C. Tracking Particular Targets*

In many application scenarios, there are only one or a few people who need to be tracked. For example, in a patient room, we only want to track the patient and ignore the movement of visitors and medical personnel. The challenge is, in order to continually track a certain person and ignore all other people in the room, the system has to first identify the person of interest. Our system can easy achieve this task by treating the problem as matching the movement of a certain tag with people moving in the room. We tested the system performance to continually track one person with one, three and five other people moving in the room. Experiments under each scenario are repeated 5 times and each experiment lasts for 5 minutes. We programmed the system to take a photo of the room and label the person with an RFID tag in the photo every second. Tracking accuracy under each scenario is defined by:

$$accuracy = \frac{\text{\# of frames correctly tracked target}}{\text{\# of frames}} \quad (5)$$

Experimental results show that the system is able to achieve similar performance for tracking a specific person compared with multi-people tracking and identification (Table II). The performance gets worse when the number of people in the room grows but the system is still able to maintain 70% tracking accuracy when there are six people in the room. The proposed tracking and identification method combines the advantage of depth sensing and passive RFID to be accurate, fast and low cost, which has a great potential to be implemented into many applications such as medical care, gaming control, and daily living monitoring.

V. DISCUSSION

*A. Limitation*

The limitations of this proposed system are generally from two aspects. First, since the proposed method uses depth information and body joints provided by the Kinect sensor, the system can only track up to six people simultaneously, and the effective working area is limited range-wise at four meters. Such a limitation is solvable by using multiple Kinect sensors or using other depth sensors to cover a larger area to track more people [25].

The second limitation is systematic since we used only two RFID antenna, if two people move in different direction along the mid-perpendicular of two antennas, the measured velocity of the two people and two tags should be the same and thus the system cannot distinguish them. The only solution to this problem is to put one more antenna in a perpendicular direction to the current antenna. However, if one extra antenna is attached to the RFID reader, the phase angle measurement speed drops and may not be able to meet the requirement to have at least two phase readings for each tag before frequency hopping. Therefore, an extra RFID reader is required but this will increase hardware cost.

Besides the limitations of the system, we also further explored the causes of identity mismatching. The errors are generally due to the velocity measurement error caused by the RFID system and the matching error caused by our matching strategy. We collected the velocity measurements during our experiments with different number of people in the room (experiments in section IV) and compared them with the velocity measured by Kinect (Ground Truth). Our experimental results show that the average error gets larger when the number of people in the room increased, but the averaged velocity measurement error for 6 people scenario (32cm/s) was not significantly increased compared with one-person scenario (21cm/s). We then analysis the velocity measurement error only for when the system made identification errors. The results show 73% of time the RFID system made large velocity measurement error (greater than 50cm/s). The rest 27% of time the system fails to make the right identification due to multiple tag and multiple person share very similar motion status such as standing still for a long time. Such results indicate that when people are standing still, the errors are most likely to be caused by matching process since multiple tag and multiple person shares very similar moving status. The potential solution will be users or other information source to distinguish stationary people. If people are in motion the errors are more likely to be caused by the RFID based velocity measurement. Such problem can be addressed by adding more antennas to provide additional velocity information.

*B. Extensions*

People tracking and identification can be applied as part of various applications, such as elder care or patient daily living monitoring. A system based on our method can track the person of interest and ignore the other people (e.g. visitors). Given the people locations and room layout mapping using Kinect [25], the system can better recognize their activities. Details of activity can be estimated using the 25-joint body skeleton provided by the Kinect sensor. Since we use only the depth sensor built into Kinect (along with passive RFID), the system can operate continuously under different lighting conditions. Using multiple Kinect sensors and RFID readers to cover larger areas for precise people tracking and identification is another potential extension that we are planning to pursue.

VI. CONCLUSIONS

We introduced a novel framework for combining passive RFID and Kinect depth sensing for accurate people tracking (centimeter level) and identification. To address the challenges in real-world application, we introduced a Doppler shift measurement approach works with frequency hopping and drop method to match the velocity measured by Kinect and RFID system. The system performance under different environment with different number of tags and different number of antennas attached to the reader is also evaluated and reported in the paper which is important for system implementation.

ACKNOWLEDGMENTS

The authors are grateful to Xuechao Pan, Sen Yang, Moliang Zhou, and Jonathan Johannemann for comments on the paper.

REFERENCES

[1] Kim, M., et al. "Face tracking and recognition with visual constraints in real-world videos." In Proc. CVPR 2008, pp: 1-8.
[2] Lu, W., et al. "A color histogram based people tracking system." In Proc. ISCAS 2001, pp: 137-140.
[3] Park, U., et al. (2013). Face tracking and recognition at a distance: A coaxial and concentric PTZ camera system. IEEE Trans. on Information Forensics and Security, 8(10), pp: 1665-1677.
[4] Bellotto, N., et al. "Multisensor data fusion for joint people tracking and identification with a service robot." In Proc. ROBIO 2007, pp: 1494-1499.
[5] Ni, R., et al. "Zoning positioning model based on minimize RFID reader." In Proc. IMSNA, 2011, pp: 117-121.
[6] Ni, L. M., et al. (2004). LANDMARC: indoor location sensing using active RFID. Wireless networks, 10(6), 701-710.
[7] Parlak, S., et al. "Non-intrusive localization of passive RFID tagged objects in an indoor workplace." In Proc. IEEE RFID-TA, 2011, pp: 181-187.
[8] Yang, L., et al. "Tagoram: Real-time tracking of mobile RFID tags to high precision using COTS devices." In Proc. ACM MobiComp, 2014, pp: 237-248.
[9] Samson, M. M., et al. (2001). Differences in gait parameters at a preferred walking speed in healthy subjects due to age, height and body weight. Aging Clinical and Experimental Research, 13(1), 16-21.
[10] Han, J., et al. "Cbid: A customer behavior identification system using passive tags." In Proc. ICNP, 201, pp: 47-58.
[11] Zhou, J., et al. "Two-dimension localization of passive RFID tags using AOA estimation." In Proc. IEEE I2MTC, 2011, pp: 1-5.
[12] Fleuret, F., et al. (2008). Multicamera people tracking with a probabilistic occupancy map. IEEE Transactions on Pattern Analysis and Machine Intelligence, 30(2), 267-282.
[13] Kim, D. Y., et al. "Data fusion in 3D vision using a RGB-D data via switching observation model and its application to people tracking." In Proc. ICCAIS, 2013, pp: 91:96.
[14] Luber, M., et al. "People tracking in RGB-d data with on-line boosted target models." In Proc. IROS, 2011, pp: 3844-3849.
[15] Haghighat, M., et al. (2013, August). Identification using encrypted biometrics. In International Conference on Computer Analysis of Images and Patterns (pp. 440-448). Springer Berlin Heidelberg.
[16] Valera, J., et al "A Review on Facial Recognition for Online Learning Authentication." In Proc. BSBT, 2015, pp: 16-19.
[17] Polívka, M., et al. (2009). UHF RF identification of people in indoor and open areas. IEEE Transactions on Microwave Theory and Techniques, 57(5), 1341-1347.
[18] D'Souza, M., et al. (2013). Evaluation of realtime people tracking for indoor environments using ubiquitous motion sensors and limited wireless network infrastructure. Pervasive and Mobile Computing, 9(4), 498-515.
[19] Zhao, Y., et al. "VIRE: Active RFID-based localization using virtual reference elimination." In Proc. ICPP, 2007, pp: 56-56.
[20] Liu, H., et al. (2007). Survey of wireless indoor positioning techniques and systems. IEEE Transactions on Systems, Man, and Cybernetics, Part C (Applications and Reviews), 37(6), 1067-1080.
[21] Germa, T., et al. (2010). Vision and RFID data fusion for tracking people in crowds by a mobile robot. Computer Vision and Image Understanding, 114(6), 641-651.
[22] Massimiliano, D., et al. (2011). Fusion of radio and video localization for people tracking. Lect. Notes Comput. Sci, 2011, 258-263.
[23] Wang, c., et al. "RFID & vision based indoor positioning and identification system." In Proc. ICCSN, 2011, pp: 506-510.
[24] Li, X., et al. (2016, May). Privacy Preserving Dynamic Room Layout Mapping. In International Conference on Image and Signal Processing (pp. 61-70). Springer International Publishing.
[25] Martínez-Zarzuela, M., et al. "Indoor 3d video monitoring using multiple kinect depth-cameras." arXiv preprint arXiv:1403.2895 (2014).
[26] Gabel, M., et al. "Full body gait analysis with Kinect." In Proc. EMBC, 2012, pp: 1964-1967.
[27] Funaya, H., et al. "Accuracy assessment of kinect body tracker in instant posturography for balance disorders." In Proc. ISMICT, 2013, pp: 213-217.
[28] Tin, T. T., et al. "Measuring Similarity between Vehicle Speed Records Using Dynamic Time Warping." In Proc. KSE, 2015, pp: 168-173.